\pdfoutput=1
\documentclass{article}
\usepackage{iclr2027_conference}
\usepackage{mathptmx}
\usepackage{amsmath,amssymb,amsthm}
\usepackage{graphicx}
\usepackage{booktabs}
\usepackage{xcolor}
\usepackage{url}
\usepackage[hidelinks]{hyperref}
\usepackage{microtype}
\usepackage{capt-of}
\usepackage{float}
\newtheorem{proposition}{Proposition}
\newcommand{\C}{\mathcal C}
\newcommand{\softmax}{\operatorname{softmax}}

\title{From Sequence Uncertainty to Phenotype Hypotheses:\\
Uncertainty-Conditioned Inference over Biomedical Knowledge Graphs}
\author{Huixin Zhan \\
New Mexico Institute of Mining and Technology \\
\And
Jason H. Moore \\
Cedars-Sinai Medical Center}

\iclrfinalcopy
\begin{document}
\maketitle
\lhead{Preprint. Under review.}
\thispagestyle{fancy}

\begin{abstract}
Genomic foundation models provide variant-level evidence, but connecting that
evidence to phenotype hypotheses requires accounting for uncertainty throughout
structured reasoning. We introduce an uncertainty-conditioned path-energy
framework that converts sequence predictions and calibrated predictive
uncertainty into explicit distributions over graph-supported phenotype
hypotheses, without patient-level phenotype supervision. An interpretable
learned energy scores admissible gene--disease--phenotype routes, query
uncertainty controls the target temperature, and path probabilities are
aggregated into phenotype hypotheses. This formulation makes evidence quality
an explicit determinant of hypothesis breadth while retaining traceable
multi-path support. Exact normalization over enumerable candidate sets provides
a reference for evaluating a Hastings-corrected learned revision operator. On
AlzKB and Hetionet, phenotype-level entropy tracks upstream uncertainty, with
Spearman correlations of 0.951 and 0.878, respectively; ablations identify the
contribution of uncertainty conditioning. In masked-edge recovery, the model
exceeds chance and degree-matched probability-permutation controls. Training the energy under the uncertainty-dependent temperature yields
higher held-out recall@10 than training under a constant temperature of
comparable magnitude. Under gene hold-out, recall@10 exceeds that of a trained
DistMult embedding on 24 of 30 graph masks (0.183 versus 0.123); this
difference is attributable mainly to three genes with many variants and is
not established for all genes. These results establish an
uncertainty-aware interface from genomic predictions to graph-supported
phenotype hypotheses that combines interpretable probability allocation
with held-out association recovery of explicitly stated scope.
\end{abstract}

\section{Introduction}
Interpreting a genetic variant requires connecting molecular evidence to the
phenotypes it could plausibly explain. Genomic sequence models provide
variant-level predictions \citep{nguyen2024sequence,brixi2026genome,zhan2025disease}, while biomedical
knowledge graphs connect genes to diseases and phenotypes through multiple
evidence routes. The quality of the upstream evidence should influence how
probability is distributed across these competing explanations: confident
evidence can support a concentrated set of hypotheses, whereas uncertain
evidence calls for a broader allocation. A point prediction alone does not
specify this allocation.

Existing work supplies complementary components.
Gene--phenotype language models answer genetics questions \citep{lyu2024gpgpt},
database-grounded agents generate and verify functional descriptions
\citep{wang2025geneagent}, and graph reasoners retrieve interpretable evidence
paths \citep{luo2024rog,sun2024tog}. We address the interface between upstream
evidence and downstream hypotheses: given a sequence prediction and its
predictive uncertainty, how should the model allocate probability across
graph-supported phenotype explanations?

We formulate this problem as uncertainty-conditioned inference over admissible
paths. Given a query gene, a sequence prediction $p$, and calibrated
uncertainty $H$, an interpretable learned utility $U_w$ defines the target
$\pi_H(\pi)\propto\exp\{U_w(\pi)/\theta(H)\}$. Aggregating path mass yields a
normalized distribution over phenotype hypotheses. The learned energy
determines the relative support for competing routes, while the
uncertainty-dependent temperature controls their concentration. Enumerable
candidate supports allow exact evaluation of the target and direct
verification of a Hastings-corrected learned revision operator
(Section~\ref{sec:sampler-scope}).

We make three contributions. \textbf{Uncertainty-conditioned phenotype
inference.} We introduce a query-conditioned probabilistic interface that
carries genomic predictive uncertainty into a structured distribution over
phenotype hypotheses, connecting evidence quality to hypothesis breadth.
\textbf{Interpretable and auditable structured inference.} We learn interpretable scores over graph routes, aggregate evidence for each phenotype, and verify the sampling procedure against the exact target distribution. \textbf{Held-out recovery with stated scope.} Across AlzKB and Hetionet, we separately evaluate uncertainty response, held-out association recovery, and sampling accuracy. Under gene hold-out, recovery exceeds chance in pooled evaluations over 40 graph masks and under disease hold-out; training under the uncertainty schedule improves recall@10 by 0.04--0.06 over constant-temperature training; and query-weighted recall@10 exceeds DistMult on 24 of 30 additional masks, an advantage concentrated in three genes. Matched controls further show that recovery depends on where the model places probability mass.

\section{Related work}
\paragraph{Genomic prediction and phenotype interpretation.}
Genomic sequence models provide variant-effect predictions
\citep{nguyen2024sequence,brixi2026genome}. Language models adapted to
genomic sequence support genome understanding and expression prediction
\citep{zhan2024efficient,honig2024long}, disease-specific language models
predict variant pathogenicity \citep{zhan2025disease}, and the underlying
architectures are reviewed by \citet{zhang2025expanding}. Gene--phenotype language models such as
GP-GPT answer genetics questions and determine relations in text
\citep{lyu2024gpgpt}, while database-grounded agents generate and verify
functional descriptions of gene sets \citep{wang2025geneagent}. Our framework
connects these levels of interpretation through an explicit probabilistic
interface: a variant-level prediction and its calibrated uncertainty determine
a distribution over graph-supported phenotype hypotheses. The central object is
the allocation of probability across competing explanations, with upstream
evidence quality controlling the breadth of that allocation.

\paragraph{Graph-supported reasoning and explanation.}
Path-based reasoners learn graph traversal policies
\citep{xiong2017deeppath,das2017go}, RoG retrieves evidence through
planned relation paths \citep{luo2024rog}, and ToG searches a graph to support
an answer \citep{sun2024tog}. Embedding methods and relational graph networks
score graph relations
\citep{bordes2013translating,yang2014embedding,schlichtkrull2018rgcn}, while
PageRank-style propagation and relation-path features provide complementary
graph-based mechanisms \citep{page1999pagerank,lao2011pra}. Our formulation
makes the query-conditioned distribution over graph-supported hypotheses an
explicit inference output, evaluated both for uncertainty response and for
recovery of masked associations. Exact normalization over admissible paths and
aggregation of multi-route evidence make this output directly inspectable.
Comparisons with DistMult and PageRank use identical candidate supports;
under gene hold-out, query-weighted recovery is higher than DistMult's on most
masks, with the scope of that comparison stated in Section~\ref{sec:c3}.

\paragraph{Probabilistic and uncertainty-aware reasoning.}
DeepProbLog integrates neural probabilities into probabilistic logic programs
\citep{manhaeve2018deepproblog}, UKGE models confidence in graph facts
\citep{chen2019ukge}, and BetaE represents logical queries with distributions
whose entropy reflects answer-set structure \citep{ren2020betae}. Agent-level
methods propagate uncertainty along reasoning trajectories
\citep{zhao2025uncertaintyagent}, while predictive uncertainty estimation,
calibration, and selective prediction characterize a predictor's reliability
\citep{gal2016dropout,lakshminarayanan2017simple,guo2017calibration,geifman2017selective};
for genomic foundation models, robustness to adversarial perturbation has
also been evaluated \citep{zhan2025safegenes}.
We use predictive uncertainty supplied by an upstream sequence model to
condition a downstream phenotype distribution. This couples uncertainty about
the input evidence to the concentration of alternative graph-supported
explanations, complementing approaches that model confidence in graph facts or
assess the reliability of a reasoning output.

\paragraph{Structured distributions and sampling.}
Finite energy models, temperature-controlled Gibbs distributions, and
Metropolis--Hastings transitions provide the probabilistic foundation for our
construction \citep{metropolis1953equation,hastings1970mc}. GFlowNets learn to
sample structured objects in proportion to reward, including
temperature-conditioned variants with controllable diversity
\citep{bengio2021gflownet,kim2024logitgfn}. We use temperature as an interface
to query-specific predictive uncertainty, making hypothesis breadth depend on
evidence quality rather than solely on a diversity setting. Enumerable path
supports additionally permit exact normalization and direct measurement of
sampling error. This separates the behavior and predictive utility of the
learned distribution from the fidelity of the operator used to sample it.

\section{Method}
\subsection{Problem setup and valid support}
Let $G=(V,E)$ be a directed typed graph and $g$ a start node. The finite support
$\C(g)$ contains complete valid paths ending in a designated outcome type.
Every consecutive node pair must be an observed graph edge or a documented
semantic inverse. For each variant, the upstream sequence model supplies a prediction
$p \in [0,1]$ calibrated using definitive Pathogenic/Benign labels,
together with a calibrated uncertainty score $H \in [0,1]$.
Here, $p$ summarizes the model's evidence about variant pathogenicity,
while higher $H$ indicates greater uncertainty in that evidence.

For AlzKB, admissible phenotype-reaching paths follow the template
\[
\mathrm{Gene}_0\rightarrow[\mathrm{Gene}_1]\rightarrow
\mathrm{Disease}\rightarrow\mathrm{Phenotype},
\]
where the gene relay is optional and uses a typed interaction, regulation, or
co-variation edge. We retain all direct routes and at most 24 deterministic,
relation-stratified route prefixes. BiologicalProcess and Pathway neighbors
annotate route genes but cannot be traversed to phenotypes in this release.

\subsection{Interpretable learned energy}
Each path has a 14-dimensional feature vector containing relation-frequency
structure, length, gene-edge type, disease coverage, endpoint specificity,
endpoint-group indicators and crowding, route-gene mechanism annotations, and
interactions with $p$ and $1-H$. Standardized features define
$U_w(\pi)=w^\top\widetilde\phi(\pi)$.
For labeled query $v$, let $c_i$ mark a prespecified core outcome and
$m_v=\sum_i c_i\pi_H(\pi_i)$. A logistic readout predicts
$\hat y_v=\sigma(\alpha m_v+\beta)$. We optimize
\[
\frac1N\sum_v\mathrm{BCE}(y_v,\hat y_v)
+\frac{\lambda}{2}\lVert w\rVert_2^2
+\frac{\lambda_\alpha}{2}\alpha^2,
\]
using bounded L-BFGS, $\lambda=10^{-2}$, and $\lambda_\alpha=10^{-3}$. Only
121 definitive Pathogenic or Benign variants \citep{richards2015acmg} enter sequence calibration or
energy learning.

\subsection{Uncertainty-conditioned target}
One schedule defines the target:
\[
\theta(H)=0.35+0.90H,\qquad
\pi_H(i)=\frac{\exp\{U_i/\theta(H)\}}{\sum_j\exp\{U_j/\theta(H)\}}.
\]
We obtain each phenotype's probability by summing the probabilities
of all paths ending at that phenotype, and group probabilities by
summing across phenotypes within each group, without applying another
softmax. These probabilities express relative model support among
graph-supported hypotheses for a given variant, not the clinical
probability that a patient has or will develop a phenotype.

\begin{proposition}[Fixed-energy entropy monotonicity]
For fixed finite utilities $U$, the Shannon entropy of $\softmax(U/\theta)$
satisfies $\frac{d\mathcal H}{d\theta}=\operatorname{Var}_{\pi_\theta}(U)/\theta^3\ge0$.
Thus any non-decreasing $\theta(H)$ produces non-decreasing path entropy at
fixed $U$.
\end{proposition}
This result characterizes path-level concentration at fixed utilities. We
evaluate phenotype-level breadth after path aggregation, including the effects
of uncertainty-dependent utility features, in the experiments below.

\subsection{Hastings-corrected learned revision}
We use the full-support independence proposal
$q_H(i)=0.90\,\softmax_i\{U/(2\theta(H))\}+0.10/|\C(g)|$.
From current state $i$, propose $j\sim q_H$ and accept with
\[
A(i,j)=\min\left\{1,\,
\exp\left(\frac{U_j-U_i}{\theta(H)}\right)\frac{q_H(i)}{q_H(j)}\right\}.
\]
\begin{proposition}[Target invariance]
The transition kernel above is irreducible, aperiodic, and reversible with
respect to $\pi_H$; consequently $\pi_H$ is stationary.
\end{proposition}
Full support follows from the uniform component, and detailed balance follows
from the standard MH ratio. We initialize at the first deterministically sorted
shortest path and run $50+10\lceil\log_2|\C(g)|\rceil+\operatorname{round}(25H)$
revisions in each of 512 independent chains. The measured mixing curve
(\S\ref{sec:fidelity}) shows the chain reaches its Monte-Carlo floor within
about ten revisions; the longer schedule is a conservative margin, not a
requirement. Main distributional statistics use the exact finite target; chain
outputs are used to test implementation fidelity and final-state coverage.

\paragraph{Exact-reference verification of learned revision.}
\label{sec:sampler-scope}
Enumerable candidate supports make the target directly computable, allowing
the revision operator to be evaluated against an exact reference. We use the
exact target for distributional statistics and the chain outputs to assess
sampling fidelity and coverage. The Hastings factor preserves the specified
uncertainty-conditioned distribution under the learned proposal
\citep{metropolis1953equation,hastings1970mc}. In particular, an independence
proposal $q_H$ combined with an uncorrected Metropolis acceptance ratio
generally has stationary mass proportional to $\pi_H(i)\,q_H(i)$ rather than
$\pi_H(i)$; the correction removes this proposal-induced distortion, which
would otherwise sharpen the distribution and undo part of the uncertainty
conditioning. The resulting operator provides a verified sampling
implementation for the formulated target. On the finite supports studied here,
direct categorical sampling is also available; the purpose of the chain
experiments is to establish fidelity, not a runtime improvement over exact
sampling. Uncapped supports grow roughly a hundredfold per additional gene
hop (App.~\ref{app:support}), the regime in which sampling replaces
enumeration and a local or amortized proposal replaces the independence
proposal used here.

\begin{figure}[t]
\centering
\includegraphics[width=.92\linewidth]{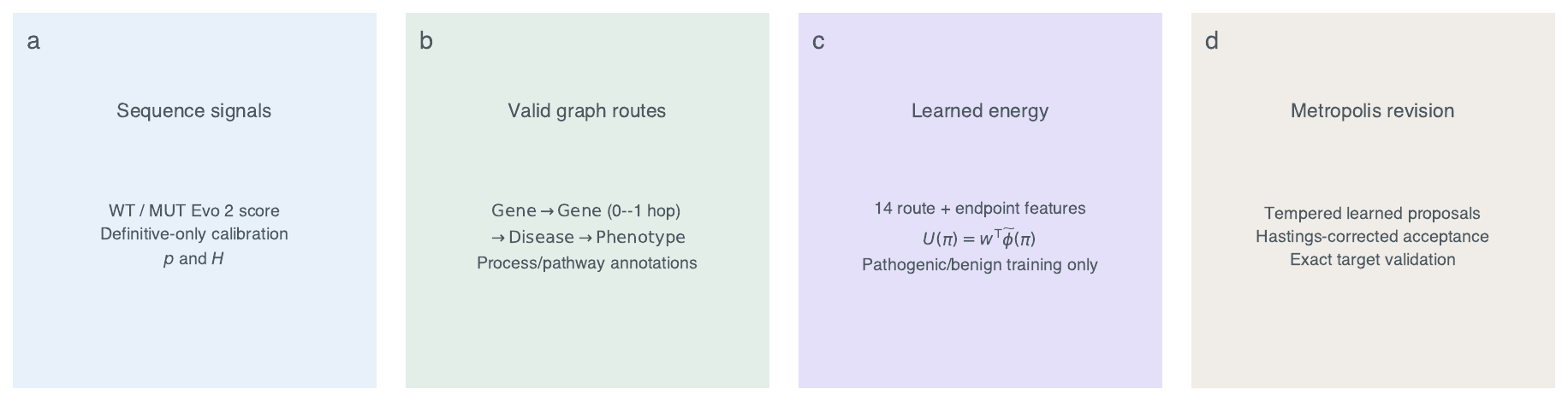}
\caption{Uncertainty-conditioned path-energy inference. Sequence predictions
and calibrated uncertainty condition an interpretable energy over valid zero-
or one-gene-hop routes. Path probabilities are aggregated into phenotype
hypotheses, and a full-support learned/uniform revision proposal is accepted
with the Hastings correction and evaluated against the exact target.}
\label{fig:framework}
\end{figure}

\section{Experimental design}
\paragraph{Case study.}
We use 985 Evo\,2-scored ClinVar \citep{landrum2018clinvar} variants matched to
AlzKB \citep{romano2024alzheimer}; 982 have a valid
path. 3 variants from GRID2IP, PLCD1, SORCS1 are excluded for absent routes.
Each query corresponds to one variant. Variants in the same gene
share the candidate graph paths, but their individual $p$ and $H$
values can produce different probability distributions over phenotypes.
The 982 variants span 15 genes, with APP, PSEN1, and PSEN2 accounting
for 94\% of queries; confidence intervals therefore resample genes
as clusters.
Candidate sets contain at most 20 route prefixes and 1060 complete paths.
Sequence probabilities and isotonic uncertainty \citep{zadrozny2002transforming} are calibrated using definitive
labels only.

\paragraph{Comparisons on identical supports.}
Component controls use the same support and learned weights: uniform path mass;
fixed temperature $\theta=0.80$ (applied at inference to the energy
trained under $\theta(H)$; energies retrained at constant temperature are in
App.~\ref{app:theta}); removal of confidence interactions; both
removals; and removal of route mechanism features. Non-learned baselines are
personalized PageRank from the start gene on the retained graph
($\alpha=0.85$, restricted to reachable endpoints and renormalized) and
beam truncation, which places uniform mass on the top-$k$ paths under the
learned utility without an uncertainty-scaled temperature ($k\in\{1,5,20\}$).
These are component controls, not competitive
baselines; the trained comparator is the DistMult reference of Section~\ref{sec:c3}.

\paragraph{Sampler checks.}
Mixing is measured by running 2,048 independent chains per query and recording
the empirical-to-exact divergence at fixed step counts (1--100), at the paper's
schedule, at twice the schedule, and at the earlier $10+\operatorname{round}(15H)$
schedule; chain-count sensitivity uses the first $N\in\{32,128,512,2048\}$
chains. The synthetic check generates heterogeneous utilities over ten groups in
spaces of 53, 256, and 1,024 states with five uncertainty levels, 30 replicates,
and 512 chains.

\paragraph{Second graph.}
Hetionet v1.0 \citep{himmelstein2017systematic} is expressed in the same schema
(34,723 nodes, 1,341,335 edges over gene, disease, symptom, process and pathway
types) so that enumeration, features, learning, target and sampler run
unchanged. Its 3,357 disease--symptom edges leave 133 diseases, and 373 of the
415 linked symptoms have several disease parents (up to 49), so endpoint
availability is gene-dependent. The same 985 variants keep their frozen
$(p,H)$ and labels; all 18 genes have valid routes (13--967 candidate paths,
13--330 endpoints). A knowledge graph without patient outcomes offers one source of ground truth:
its own edges. We therefore test predictive generalization by masking 20\% of
disease--symptom edges (671 of 3,357), rebuilding supports and retraining the
energy on the reduced graph, and asking whether the path distribution places
mass on true outcomes it never saw---held-out symptoms of each gene's own
diseases that remain reachable on the reduced graph (recall@$k$, MRR; 964
variants in 15 genes). The ranking borrows the filtered protocol of
knowledge-graph completion, but the object evaluated is the query-conditioned
path distribution, not a triple scorer. Filtering removes every symptom still
attached to the gene's own diseases, so a recoverable held-out symptom is
reached only through gene relays $g\to g'\to d'\to s$ (worked query:
Appendix~\ref{app:example}). A trained DistMult
embedding \citep{yang2014embedding} (dimension 32, logistic loss, Adagrad,
two-hop composition) is the reference for a trained triple scorer on the same
candidates; it receives no $(p,H)$. Two prespecified tests follow, with recall@10
primary. \emph{Matched controls} keep each query's candidates, positives and
full-model probabilities and compare against chance, symptom popularity
(degree in the reduced graph), and a degree-matched permutation that shuffles
the model's probabilities among candidates of similar degree (one permutation
per gene, 200 draws), preserving values and entropy but not their assignment. Variants in the same
gene share the same reachable held-out symptoms; recovery metrics average
over variant queries unless labeled gene-balanced (equal weight per gene).
Gene-held-out evaluation excludes all variants in one group
(APP, PSEN1, PSEN2, or the remaining genes) from calibration and
energy fitting, using graph-mask seeds 42, 7, and 123.
The held-out genes remain in the graph used for inference.
A noninferiority margin of 0.02 against DistMult was fixed in advance; leakage
and coverage checks are in App.~\ref{app:checks}. Additional evaluations
(App.~\ref{app:add}) comprise seven further masks under the same protocol; 30
additional masks under a protocol fixed before they were run, with DistMult
averaged over three training seeds; schedule sensitivity with the energy
retrained; component ablations on the recovery task; corrupted uncertainty;
and disease hold-out.

\paragraph{Energy-recovery study.}
Queries are generated with $K$ candidate paths, 14-dimensional standardized
features including a binary core-outcome indicator, a planted energy $w^\ast$,
and labels drawn from the same readout family as the model. The exact training
routine of the paper is applied for $K\in\{53,256,1024\}$ and
$N\in\{121,500,2000\}$ over five replicates. Recovery is measured on 500
held-out queries by weight cosine, utility rank agreement, group-level JS
between learned and planted targets, and label AUROC against the oracle readout.

\section{Results}
\subsection{The implemented chain samples the stated target}
\label{sec:fidelity}
\begin{figure}[t]
\centering
\includegraphics[width=.74\linewidth]{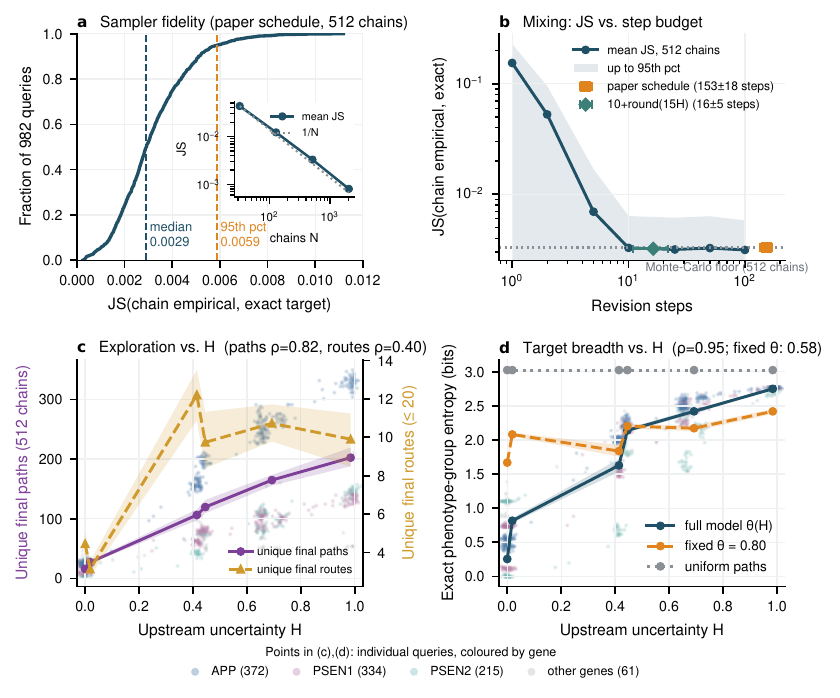}
\caption{Sampler fidelity and uncertainty behavior on 982 queries (AlzKB). (a)
Empirical cumulative distribution of chain-to-target JS at the paper's
schedule; inset: mean JS versus number of chains with a $1/N$ reference. (b)
Mixing curve: mean JS (shaded to the 95th percentile) versus revision steps;
the paper's and the earlier schedule are marked (legend top right). (c)
Unique final paths (left axis, purple) and routes (right axis, gold, capped at
20) across 512 chains versus $H$; gene-cluster 95\% CIs 0.79--0.96 (paths),
0.10--0.78 (routes). (d) Exact phenotype-group entropy versus $H$ for the full
model, fixed $\theta$, and uniform paths (legend bottom right); cluster CIs
0.90--0.98 (full), 0.47--0.85 (fixed $\theta$). Curves are quantile-binned
means with 95\% bootstrap bands. Small points in (c),(d) are individual
queries coloured by gene (key with counts below the panels), jittered in $H$
because the isotonic calibrator is a step function; the vertical spread
within an $H$ plateau is gene-dependent support.}
\label{fig:mainresults}
\end{figure}

Across 982 real queries, the sampled phenotype-group distribution closely
matches the exact target, with mean Jensen--Shannon divergence (JS)
\citep{lin1991jsd} of 0.0031 (median 0.0029) and mean acceptance rate
of 0.507 (Fig.~\ref{fig:mainresults}a). Increasing the number of chains
from 32 to 128, 512, and 2,048 reduces JS from 0.043 to 0.012, 0.0033,
and 0.0008, respectively. This approximately $1/N$ decrease indicates
that the remaining difference is consistent with finite-sample error
rather than persistent sampling bias. Convergence is also rapid:
JS falls from 0.155 after one revision to 0.053, 0.0070, and 0.0034
after two, five, and ten revisions, then levels off
(Fig.~\ref{fig:mainresults}b); both the main schedule (mean 153 steps) and
the shorter one (mean 16 steps) sit on this floor. Synthetic experiments show similar agreement
across uncertainty levels and state spaces of up to 1,024 states:
the largest condition-level mean JS is 0.0038 and the largest
95th-percentile JS is 0.0069, with little variation across $H$ or
state-space size (Fig.~\ref{fig:synthetic}a). Together, these results
verify the sampling implementation and distinguish finite-sample error
from the behavior of the target distribution, which we evaluate next.

\subsection{Upstream uncertainty controls structured hypothesis breadth}
Phenotype hypotheses broaden as uncertainty increases. From the lowest to
the highest uncertainty quantile, phenotype-group entropy rises from 0.585
to 2.631 bits, while the probability assigned to the top-ranked group falls
from 0.864 to 0.357. Across variants, entropy correlates strongly with $H$
at both the group level ($\rho=0.951$, gene-cluster 95\% CI 0.898--0.984)
and the individual-phenotype level ($\rho=0.890$, 0.852--0.926). Final
states from 512 chains also become more diverse with higher $H$, with
correlations of $\rho=0.821$ (0.792--0.961) for paths and $\rho=0.403$
(0.097--0.775) for routes. The cap of 20 routes per gene limits
route-level variation (Fig.~\ref{fig:mainresults}c,d). Recomputing the
target with $k$NN-calibrated uncertainty instead of isotonic calibration
preserves a strong entropy--uncertainty correlation ($\rho=0.869$,
0.777--0.964), showing that the relationship holds under both calibrators.
The association is not driven by the three high-count genes: without APP,
PSEN1, and PSEN2 it is $\rho=0.950$ (61 variants, 12 genes), and the median
within-gene correlation is 0.959 (App.~\ref{app:genes}).
The following ablations examine how explicit uncertainty conditioning
contributes to this relationship, while the masked-edge experiments test
whether the resulting probability assignments help recover held-out
associations.

\subsection{Ablations separate breadth control from probability allocation}
\begin{figure}[t]
\centering
\includegraphics[width=.76\linewidth]{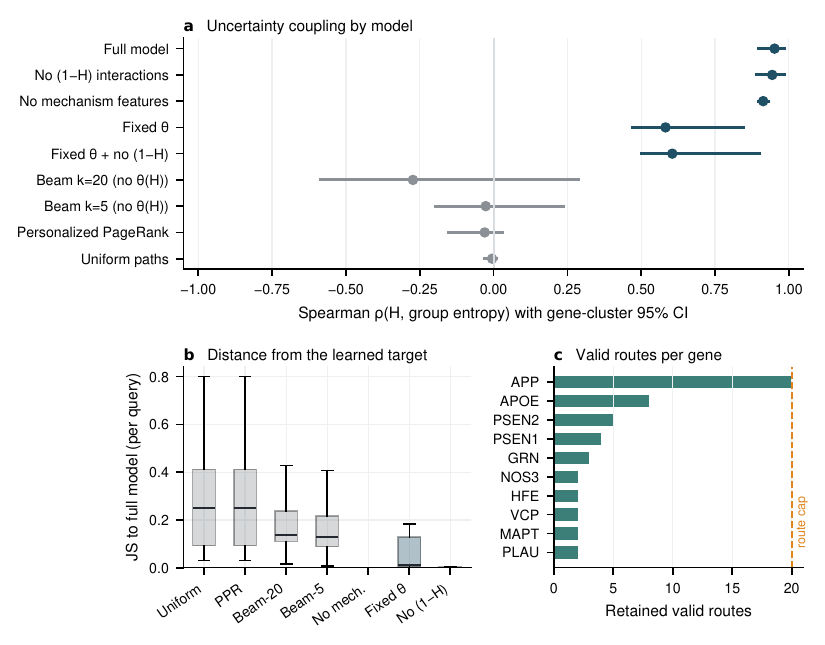}
\caption{Controls and baselines (AlzKB). (a) $H$--entropy association with
gene-cluster 95\% CIs; navy points are component controls, grey points are
non-learned or truncation baselines (no separate legend; the row labels name
each model). (b) Per-query JS to the full model, one box per model. (c)
Retained valid routes per gene; the dashed line marks the route cap of 20.
Every route reaches the same 53 endpoints through one disease node, so
uniform and PageRank propagation coincide.}
\label{fig:ablations}
\end{figure}

Table~\ref{tab:ablations} and Fig.~\ref{fig:ablations} separate the effects of
uncertainty conditioning and learned route features. Fixing the temperature
at inference reduces the correlation between upstream uncertainty and phenotype-group
entropy from 0.951 to 0.582, and removing both the temperature schedule and
confidence interactions gives 0.605 (intervals in Table~\ref{tab:ablations}):
these controls attenuate the strong uncertainty--breadth relationship of the
full model. Removing mechanism
features changes the target distribution (mean JS 0.073) while retaining a
high entropy correlation, showing that route annotations affect probability
allocation beyond the temperature schedule. Uniform and personalized PageRank
propagation exhibit near-zero associations with uncertainty, and beam
truncation under the learned utility does not reproduce the full model's
relationship. Together, the controls distinguish the uncertainty-dependent
breadth of the target from the route features that shape its allocation.
With the energy retrained, breadth and recovery are stable across
schedules $a+bH$ with $b>0$, whereas constant-temperature energies recover
less (App.~\ref{app:theta}).

\begin{table}[t]
\caption{Component controls and baselines on identical candidate supports
(982 queries). Spearman $\rho$ is between $H$ and grouped entropy; brackets are
gene-cluster 95\% bootstrap intervals over 15 genes.}
\label{tab:ablations}
\centering
\small
\begin{tabular}{lrrrl}
\toprule
Model & Entropy & Top-1 & JS to full & $\rho$ [95\% CI] \\
\midrule
Uniform paths & 3.025 & 0.264 & 0.287 & $-0.005$ [$-0.03$, $0.01$] \\
Personalized PageRank & 3.025 & 0.264 & 0.287 & $-0.030$ [$-0.15$, $0.03$] \\
Beam $k=20$ (no $\theta(H)$) & 2.060 & 0.414 & 0.186 & $-0.273$ [$-0.58$, $0.28$] \\
Beam $k=5$ (no $\theta(H)$) & 0.840 & 0.726 & 0.158 & $-0.027$ [$-0.20$, $0.23$] \\
\midrule
Fixed $\theta$ + no confidence interactions & 2.040 & 0.508 & 0.054 & $0.605$ [$0.50$, $0.90$] \\
Fixed temperature $\theta=0.80$ & 2.078 & 0.515 & 0.052 & $0.582$ [$0.47$, $0.85$] \\
No mechanism features & 2.127 & 0.458 & 0.073 & $0.913$ [$0.90$, $0.93$] \\
No confidence interactions & 1.695 & 0.589 & 0.002 & $0.944$ [$0.89$, $0.99$] \\
Full model (reference) & 1.662 & 0.606 & 0.000 & $0.951$ [$0.90$, $0.98$] \\
\bottomrule
\end{tabular}
\end{table}

\subsection{Usable graph topology shapes hypothesis support}
AlzKB contains 210,656 retained nodes and 1,199,652 edges, but its
phenotype-reaching subgraph has a specific structure: all 53
\texttt{DISEASEMANIFESTSSYMPTOM} edges originate from one disease node, and
Pathway and BiologicalProcess nodes provide annotations rather than additional
phenotype-reaching continuations. Consequently, every evaluated gene has the
same 53 phenotype endpoints, although the routes supporting those endpoints
differ. Under the evaluated propagation baselines, personalized PageRank and
uniform path mass coincide (Table~\ref{tab:ablations}). The learned energy can
nevertheless assign different weights to the valid evidence routes. Hetionet
(Section~\ref{sec:hetionet}) provides a complementary evaluation in which
multiple disease parents and gene-dependent endpoint availability test the
same framework on a different reasoning topology. Perturbation sensitivity of the target is reported in App.~\ref{app:perturb}.

\subsection{Energy recovery and optimization diagnostics}
\label{sec:recovery}
The synthetic study tests whether the model can recover a known path
utility used to generate the data. With at most 256 candidate paths and
at least 500 training queries, successful fits achieve weight cosine
similarity of 0.98--0.99 and held-out AUROC within 0.01 of the oracle
(Fig.~\ref{fig:synthetic}d). On both real graphs, repeated fits produce
identical objective values and weight cosine similarity of 1.000 across
the tested restarts and tolerance settings. Some synthetic fits collapse
when labeled data are limited or candidate sets are large.
App.~\ref{app:synthetic} provides the full recovery results, failure
rates, and analysis of initialization choices.

\subsection{Uncertainty-conditioned breadth replicates across graph topologies}
\label{sec:hetionet}
On Hetionet, the energy trains reproducibly using the same 121 definitive
anchors, reaching objective 0.238 compared with 0.640 for a constant predictor
(App.~\ref{app:synthetic}). Endpoint entropy
of the exact target increases from 2.020 to 6.413 bits across uncertainty
strata while top-1 mass decreases from 0.529 to 0.052. The
uncertainty--entropy correlation is 0.878 (gene-cluster 95\% CI 0.780--0.910),
compared with 0.495 (0.281--0.797) under fixed temperature
(Table~\ref{tab:hetionet}, Fig.~\ref{fig:hetionet}b). Unlike AlzKB, Hetionet
yields distinct uniform and PageRank distributions, but neither exhibits a
meaningful association with uncertainty ($\rho=0.027$, $0.028$). Final-state
diversity also tracks uncertainty ($\rho=0.896$), and the sampler achieves mean
grouped JS of approximately 0.003 on supports averaging 713 states
(Fig.~\ref{fig:hetionet}c; further statistics in App.~\ref{app:hetionet}). These
results reproduce uncertainty-conditioned hypothesis breadth under
gene-dependent phenotype availability; without the three high-count
genes the correlation is 0.769 (64 variants, 15 genes; App.~\ref{app:genes}).

\subsection{Held-out recovery: evidence and scope}
\label{sec:c3}
\begin{figure}[t]
\begin{minipage}[t]{.44\linewidth}
\vspace{0pt}\centering
\includegraphics[width=\linewidth]{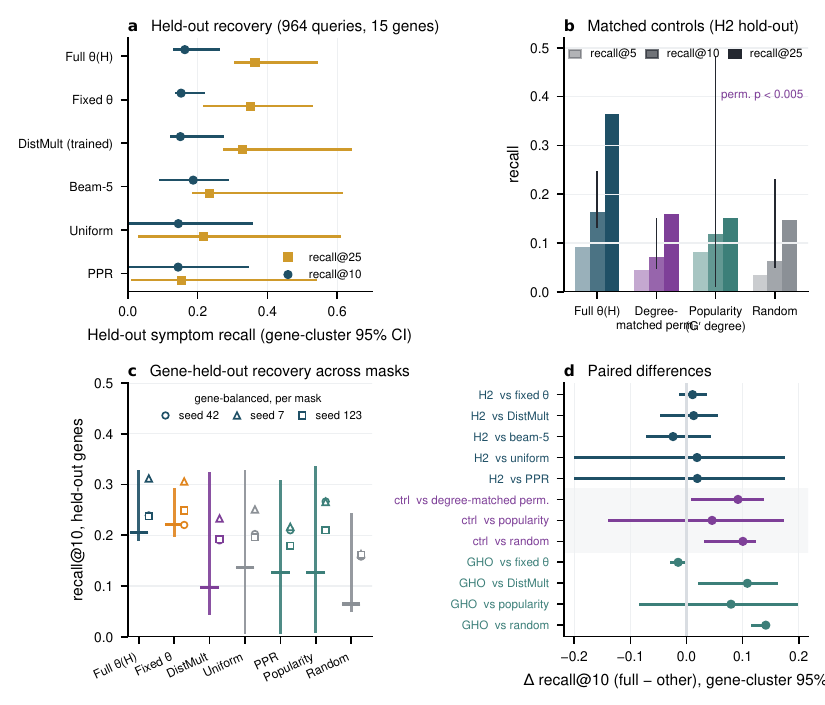}
\caption{Masked-edge recovery (Hetionet). (a) Recall of masked symptoms on
the primary hold-out, recall@10 (navy) and recall@25 (gold), gene-cluster 95\%
CIs. (b) Matched controls on identical candidates and positives; shade
encodes recall@5/10/25, bars are recall@10 CIs. (c) Gene-held-out recovery:
pooled mean with gene-cluster 95\% CI over three masks; open markers are
gene-balanced per-mask means. (d) Paired recall@10 differences, full minus
each alternative: primary hold-out (navy), matched controls (purple), gene
hold-out (teal).}
\label{fig:c3}
\end{minipage}\hfill
\begin{minipage}[t]{.54\linewidth}
\vspace{0pt}\centering\footnotesize\setlength{\tabcolsep}{3.5pt}
\captionof{table}{Gene-held-out recall@10 (Hetionet; query genes withheld
from calibration and energy learning) on the three prespecified masks and on
30 additional masks (DistMult averaged over three training seeds). QW:
query-weighted; GB: gene-balanced; brackets: gene-cluster 95\% CIs; CV:
across-mask coefficient of variation.}
\label{tab:gho}
{\scriptsize\setlength{\tabcolsep}{2.5pt}%
\begin{tabular}{@{}lccc@{}}
\toprule
 & Full, $\theta(H)$ & DistMult & Full $-$ DistMult [CI] \\
\midrule
3 masks, QW & 0.206 & 0.097 & $+0.108$ [$0.024$, $0.160$] \\
\quad per-mask range & 0.182--0.222 & 0.057--0.151 & \\
30 masks, QW & 0.183 & 0.123 & $+0.060$ [$-0.028$, $0.093$] \\
30 masks, GB & 0.252 & 0.239 & $+0.013$ [$-0.043$, $0.082$] \\
\quad masks full higher & \multicolumn{3}{c@{}}{24/30 (QW), 19/30 (GB)} \\
\quad CV, QW / GB & 0.30 / 0.16 & 0.39 / 0.12 & \\
\bottomrule
\end{tabular}}
\vspace{6pt}
\captionof{table}{Hetionet (985 queries, 18 genes; masked-edge recovery on 964
queries, 15 genes). $\rho$ is between $H$ and endpoint entropy (gene-cluster
95\% CI) and supports the breadth claim; recall and MRR are conditional on
reachable masked symptoms (primary split, Section~\ref{sec:c3}).}
\label{tab:hetionet}
{\scriptsize\setlength{\tabcolsep}{2.5pt}%
\begin{tabular}{@{}llrrr@{}}
\toprule
Model & $\rho$ [CI] & r@10 & r@25 & MRR \\
\midrule
Full model, $\theta(H)$ & $0.878$ [$0.78$, $0.91$] & 0.163 & 0.364 & 0.067 \\
Fixed temperature & $0.495$ [$0.28$, $0.80$] & 0.153 & 0.351 & 0.065 \\
DistMult (trained) & -- & 0.151 & 0.329 & 0.061 \\
Beam $k=5$ & $0.014$ [$-0.07$, $0.10$] & 0.187 & 0.235 & 0.050 \\
Uniform paths & $0.027$ [$-0.11$, $0.09$] & 0.145 & 0.217 & 0.077 \\
PageRank (personalized) & $0.028$ [$-0.12$, $0.09$] & 0.144 & 0.154 & 0.054 \\
\bottomrule
\end{tabular}
}
\end{minipage}
\end{figure}

\textbf{Gene-held-out recovery.} With each query gene excluded from sequence
calibration and energy learning, the framework achieves mean recall@10 of
0.206 across three prespecified edge masks (Table~\ref{tab:gho}, Fig.~\ref{fig:c3}c). The
paired improvement over DistMult is 0.108 (gene-cluster 95\% CI 0.024--0.160)
on these masks. The observed mean
improvement is positive on all three masks: 0.032 for seed 42 (95\% CI
$-0.008$ to $0.041$), 0.156 for seed 7 ($0.012$--$0.264$), and 0.137 for seed
123 ($0.037$--$0.227$).

\textbf{Evaluation on additional masks.} The evaluation was repeated on
seven further masks and on 30 additional masks. For the latter, the protocol
was specified before evaluation, and DistMult was trained with three seeds per
mask with its metrics averaged (App.~\ref{app:masks}, \ref{app:conf}). The
difference on the additional masks is smaller than on the three prespecified
masks. On the 30
additional masks, query-weighted recall@10 is 0.183 versus 0.123 for DistMult
and is higher on 24 of 30 masks. Gene-balanced results, intervals, per-gene results, across-mask variability,
and comparisons with chance and under disease hold-out are reported in
Table~\ref{tab:gho} and App.~\ref{app:conf} and~\ref{app:disease}.

\textbf{Recovery beyond matched controls.} On the primary masked graph,
recall@10 is 0.163, exceeding chance at 0.063 (paired difference 0.100, 95\% CI
$0.034$--$0.122$) and a degree-matched permutation of the model's probabilities
at 0.072 (paired difference 0.092, 95\% CI $0.011$--$0.136$; $p<0.005$;
Fig.~\ref{fig:c3}b,d). The permutation preserves the candidate set, positives,
probability values, and entropy while changing their assignment. This
comparison shows that recovery depends on where the model places probability
mass, rather than on distributional breadth alone.

\textbf{Primary split and conditioning ablations.} On the primary split, the
full model achieves recall@10/recall@25/MRR of 0.163/0.364/0.067, compared
with 0.151/0.329/0.061 for DistMult (Table~\ref{tab:hetionet},
Fig.~\ref{fig:c3}a). These point estimates favor the proposed method; the
paired intervals on that individual split include zero ($+0.013$, CI $-0.044$
to $0.054$). Fixed-temperature ablations applied at inference attain similar recall, with
full-minus-fixed recall@10 differences of 0.011 (95\% CI $-0.011$ to $0.033$)
on the primary split and $-0.015$ ($-0.026$ to $0.001$) under gene hold-out.
The effect of the schedule on recovery appears when it is applied
during training: an energy trained under $\theta(H)$ attains recall@10 of
0.167 on ten masks, compared with 0.106--0.124 for energies trained at
constant temperatures between 0.20 and 0.80 (paired gains 0.043--0.061, all
intervals excluding zero); the gain is smaller at constant temperatures of
1.00 and 1.25 (0.022 and 0.015, intervals including zero;
App.~\ref{app:theta}). Component ablations show that the sequence-prediction
features contribute to recovery, and hypothesis breadth follows the supplied
$H$ (App.~\ref{app:abl}, \ref{app:hpert}). Recovery is evaluated over reachable masked symptoms
on matched candidate supports; coverage, per-gene and popularity comparisons,
and the prespecified single-split noninferiority analysis are in
App.~\ref{app:checks}, and the descriptive recall--$H$ association in
App.~\ref{app:recallH}.

\section{Secondary diagnostic: variant classification}
\label{sec:diag}
Variant classification evaluates a different output from phenotype-hypothesis
recovery. Across five outer folds, graph-augmented AUROC is 0.968 compared
with 0.955 for sequence-only features, with a paired difference of 0.0126
(gene-cluster 95\% CI $-0.0235$ to $0.0191$). Brier score and expected
calibration error are higher with graph features. We report this diagnostic
separately from the recovery results; full metrics and the locked-label
protocol appear in App.~\ref{app:diag}.

\section{Future work}
Future work will extend the evaluation to more genes and biomedical
graphs, which is needed to establish recovery across genes, compare with relational and path-policy reasoners, and improve
learning with larger candidate sets and limited labeled data. Patient-level
studies could establish how the output distributions relate to observed
phenotypes and investigate their calibration for clinical risk estimation.

\section{Conclusion}
We introduced an uncertainty-conditioned path-energy framework that connects
genomic sequence predictions to explicit distributions over graph-supported
phenotype hypotheses. The framework combines interpretable route utilities,
query-dependent control of hypothesis breadth, and exact-reference validation
of learned revision. Across two biomedical knowledge graphs, hypothesis breadth
tracks upstream uncertainty, and masked-edge tests show that the learned
probability allocation recovers held-out associations beyond chance and
matched permutation controls. Training the energy under the uncertainty-dependent temperature yields
higher held-out recall@10 than training under a constant temperature of
comparable magnitude. Under gene hold-out, recall@10 exceeds that of a trained
DistMult embedding on 24 of 30 graph masks (0.183 versus 0.123); this
difference is attributable mainly to three genes with many variants and is
not established for all genes. Together, these
results advance uncertainty-aware genotype-to-phenotype interpretation through
interpretable inference and held-out association recovery of explicitly
stated scope.

\section*{Reproducibility statement}
All reported numbers are generated from frozen experiment tables and checked
programmatically against the manuscript. Split manifests, model hashes, and
per-experiment logs document the calibration, training, and evaluation
pipeline. Reproducibility checks cover path validity, feature dimensions,
detailed balance, empirical stationarity, full proposal support, empty and
singleton supports, deterministic seeds, and mass conservation. Code,
frozen result tables, figure scripts, and the manuscript-number verifier are
available online.\footnote{\url{https://github.com/huixin-zhan-ai/Uncertainty-Conditioned-Inference}}

\section*{AI use statement}
AI assistants were used to polish the writing, to audit the writing for
consistency between the reported numbers, tables, and claims, and to audit
the code. The authors reviewed all AI-assisted content, verified every
number and claim against the frozen experiment outputs, and take full
responsibility for the final manuscript and code.

\bibliography{references}

@article{brixi2026genome,
  title={Genome modelling and design across all domains of life with Evo 2},
  author={Brixi, Garyk and Durrant, Matthew G and Ku, Jerome and Naghipourfar, Mohsen and Poli, Michael and Sun, Gwanggyu and Brockman, Greg and Chang, Daniel and Fanton, Alison and Gonzalez, Gabriel A and others},
  journal={Nature},
  volume={652},
  number={8112},
  pages={1349--1361},
  year={2026},
  publisher={Nature Publishing Group UK London}
}

@article{nguyen2024sequence,
  title={Sequence modeling and design from molecular to genome scale with Evo},
  author={Nguyen, Eric and Poli, Michael and Durrant, Matthew G and Kang, Brian and Katrekar, Dhruva and Li, David B and Bartie, Liam J and Thomas, Armin W and King, Samuel H and Brixi, Garyk and others},
  journal={Science},
  volume={386},
  number={6723},
  pages={eado9336},
  year={2024},
  publisher={American Association for the Advancement of Science}
}

@article{romano2024alzheimer,
  title={The Alzheimer’s knowledge base: a knowledge graph for Alzheimer disease research},
  author={Romano, Joseph D and Truong, Van and Kumar, Rachit and Venkatesan, Mythreye and Graham, Britney E and Hao, Yun and Matsumoto, Nick and Li, Xi and Wang, Zhiping and Ritchie, Marylyn D and others},
  journal={Journal of Medical Internet Research},
  volume={26},
  pages={e46777},
  year={2024},
  publisher={JMIR Publications Toronto, Canada}
}

@article{landrum2018clinvar,
  title={ClinVar: improving access to variant interpretations and supporting evidence},
  author={Landrum, Melissa J and Lee, Jennifer M and Benson, Mark and Brown, Garth R and Chao, Chen and Chitipiralla, Shanmuga and Gu, Baoshan and Hart, Jennifer and Hoffman, Douglas and Jang, Wonhee and others},
  journal={Nucleic acids research},
  volume={46},
  number={D1},
  pages={D1062--D1067},
  year={2018},
  publisher={Oxford University Press}
}

@inproceedings{gal2016dropout,
  title={Dropout as a bayesian approximation: Representing model uncertainty in deep learning},
  author={Gal, Yarin and Ghahramani, Zoubin},
  booktitle={international conference on machine learning},
  pages={1050--1059},
  year={2016},
  organization={PMLR}
}

@inproceedings{zadrozny2002transforming,
  title={Transforming classifier scores into accurate multiclass probability estimates},
  author={Zadrozny, Bianca and Elkan, Charles},
  booktitle={Proceedings of the eighth ACM SIGKDD international conference on Knowledge discovery and data mining},
  pages={694--699},
  year={2002}
}

@article{richards2015acmg,
  title={Standards and guidelines for the interpretation of sequence variants: a joint consensus recommendation of the American College of Medical Genetics and Genomics and the Association for Molecular Pathology},
  author={Richards, Sue and Aziz, Nazneen and Bale, Sherri and Bick, David and Das, Soma and Gastier-Foster, Julie and Grody, Wayne W and Hegde, Madhuri and Lyon, Elaine and Spector, Elaine and others},
  journal={Genetics in medicine},
  volume={17},
  number={5},
  pages={405--423},
  year={2015},
  publisher={Nature Publishing Group}
}

@article{metropolis1953equation,
  title={Equation of state calculations by fast computing machines},
  author={Metropolis, Nicholas and Rosenbluth, Arianna W and Rosenbluth, Marshall N and Teller, Augusta H and Teller, Edward},
  journal={The journal of chemical physics},
  volume={21},
  number={6},
  pages={1087--1092},
  year={1953},
  publisher={American Institute of Physics}
}

@article{lakshminarayanan2017simple,
  title={Simple and scalable predictive uncertainty estimation using deep ensembles},
  author={Lakshminarayanan, Balaji and Pritzel, Alexander and Blundell, Charles},
  journal={Advances in neural information processing systems},
  volume={30},
  year={2017}
}

@inproceedings{guo2017calibration,
  title={On calibration of modern neural networks},
  author={Guo, Chuan and Pleiss, Geoff and Sun, Yu and Weinberger, Kilian Q},
  booktitle={International conference on machine learning},
  pages={1321--1330},
  year={2017},
  organization={PMLR}
}

@article{das2017go,
  title={Go for a walk and arrive at the answer: Reasoning over paths in knowledge bases using reinforcement learning},
  author={Das, Rajarshi and Dhuliawala, Shehzaad and Zaheer, Manzil and Vilnis, Luke and Durugkar, Ishan and Krishnamurthy, Akshay and Smola, Alex and McCallum, Andrew},
  journal={arXiv preprint arXiv:1711.05851},
  year={2017}
}

@inproceedings{xiong2017deeppath,
  title={Deeppath: A reinforcement learning method for knowledge graph reasoning},
  author={Xiong, Wenhan and Hoang, Thien and Wang, William Yang},
  booktitle={Proceedings of the 2017 conference on empirical methods in natural language processing},
  pages={564--573},
  year={2017}
}

@article{bordes2013translating,
  title={Translating embeddings for modeling multi-relational data},
  author={Bordes, Antoine and Usunier, Nicolas and Garcia-Duran, Alberto and Weston, Jason and Yakhnenko, Oksana},
  journal={Advances in neural information processing systems},
  volume={26},
  year={2013}
}

@inproceedings{schlichtkrull2018rgcn,
  title={Modeling relational data with graph convolutional networks},
  author={Schlichtkrull, Michael and Kipf, Thomas N and Bloem, Peter and Van Den Berg, Rianne and Titov, Ivan and Welling, Max},
  booktitle={European semantic web conference},
  pages={593--607},
  year={2018},
  organization={Springer}
}

@techreport{page1999pagerank,
  title={The {PageRank} citation ranking: Bringing order to the web},
  author={Page, Lawrence and Brin, Sergey and Motwani, Rajeev and Winograd, Terry},
  institution={Stanford InfoLab},
  year={1999}
}

@article{hastings1970mc,
  title={Monte Carlo sampling methods using Markov chains and their applications},
  author={Hastings, W Keith},
  year={1970},
  publisher={Oxford University Press}
}

@article{geifman2017selective,
  title={Selective classification for deep neural networks},
  author={Geifman, Yonatan and El-Yaniv, Ran},
  journal={Advances in neural information processing systems},
  volume={30},
  year={2017}
}

@article{lin1991jsd,
  title={Divergence measures based on the Shannon entropy},
  author={Lin, Jianhua},
  journal={IEEE Transactions on Information theory},
  volume={37},
  number={1},
  pages={145--151},
  year={1991},
  publisher={IEEE}
}

@article{himmelstein2017systematic,
  title={Systematic integration of biomedical knowledge prioritizes drugs for repurposing},
  author={Himmelstein, Daniel Scott and Lizee, Antoine and Hessler, Christine and Brueggeman, Leo and Chen, Sabrina L and Hadley, Dexter and Green, Ari and Khankhanian, Pouya and Baranzini, Sergio E},
  journal={elife},
  volume={6},
  pages={e26726},
  year={2017},
  publisher={eLife Sciences Publications, Ltd}
}

@article{yang2014embedding,
  title={Embedding entities and relations for learning and inference in knowledge bases},
  author={Yang, Bishan and Yih, Wen-tau and He, Xiaodong and Gao, Jianfeng and Deng, Li},
  journal={arXiv preprint arXiv:1412.6575},
  year={2014}
}

@article{lyu2024gpgpt,
  title={Gp-gpt: Large language model for gene-phenotype mapping},
  author={Lyu, Yanjun and Wu, Zihao and Zhang, Lu and Zhang, Jing and Li, Yiwei and Ruan, Wei and Liu, Zhengliang and Zhang, Zeyu and Li, Xiang and Liu, Rongjie and others},
  journal={arXiv preprint arXiv:2409.09825},
  year={2024}
}

@article{wang2025geneagent,
  title={GeneAgent: self-verification language agent for gene-set analysis using domain databases},
  author={Wang, Zhizheng and Jin, Qiao and Wei, Chih-Hsuan and Tian, Shubo and Lai, Po-Ting and Zhu, Qingqing and Day, Chi-Ping and Ross, Christina and Leaman, Robert and Lu, Zhiyong},
  journal={Nature Methods},
  volume={22},
  number={8},
  pages={1677--1685},
  year={2025},
  publisher={Nature Publishing Group US New York}
}

@inproceedings{luo2024rog,
  title={Reasoning on graphs: Faithful and interpretable large language model reasoning},
  author={Luo, Linhao and Li, Yuan-Fang and Haffari, Reza and Pan, Shirui},
  booktitle={International Conference on Learning Representations},
  volume={2024},
  pages={14400--14423},
  year={2024}
}

@inproceedings{sun2024tog,
  title={Think-on-graph: Deep and responsible reasoning of large language model on knowledge graph},
  author={Sun, Jiashuo and Xu, Chengjin and Tang, Lumingyuan and Wang, Saizhuo and Lin, Chen and Gong, Yeyun and Ni, Lionel and Shum, Heung-Yeung and Guo, Jian},
  booktitle={International Conference on Learning Representations},
  volume={2024},
  pages={3868--3898},
  year={2024}
}

@article{manhaeve2018deepproblog,
  title={Deepproblog: Neural probabilistic logic programming},
  author={Manhaeve, Robin and Dumancic, Sebastijan and Kimmig, Angelika and Demeester, Thomas and De Raedt, Luc},
  journal={Advances in neural information processing systems},
  volume={31},
  year={2018}
}

@inproceedings{chen2019ukge,
  title={Embedding uncertain knowledge graphs},
  author={Chen, Xuelu and Chen, Muhao and Shi, Weijia and Sun, Yizhou and Zaniolo, Carlo},
  booktitle={Proceedings of the AAAI conference on artificial intelligence},
  volume={33},
  number={01},
  pages={3363--3370},
  year={2019}
}

@article{ren2020betae,
  title={Beta embeddings for multi-hop logical reasoning in knowledge graphs},
  author={Ren, Hongyu and Leskovec, Jure},
  journal={Advances in Neural Information Processing Systems},
  volume={33},
  pages={19716--19726},
  year={2020}
}

@inproceedings{kim2024logitgfn,
  title={Learning to Scale Logits for Temperature-Conditional {GFlowNets}},
  author={Kim, Minsu and Ko, Joohwan and Yun, Taeyoung and Zhang, Dinghuai and Pan, Ling and Kim, Woo Chang and Park, Jinkyoo and Bengio, Emmanuel and Bengio, Yoshua},
  booktitle={International Conference on Machine Learning (ICML)},
  pages={24248--24270},
  year={2024}
}

@inproceedings{bengio2021gflownet,
  title={Flow Network based Generative Models for Non-Iterative Diverse Candidate Generation},
  author={Bengio, Emmanuel and Jain, Moksh and Korablyov, Maksym and Precup, Doina and Bengio, Yoshua},
  booktitle={Advances in Neural Information Processing Systems (NeurIPS)},
  year={2021}
}

@inproceedings{zhao2025uncertaintyagent,
  title={Uncertainty Propagation on {LLM} Agent},
  author={Zhao, Qiwei and Li, Dong and Liu, Yanchi and Cheng, Wei and Sun, Yiyou and Oishi, Mika and Osaki, Takao and Matsuda, Katsushi and Yao, Huaxiu and Zhao, Chen and Chen, Haifeng and Zhao, Xujiang},
  booktitle={Annual Meeting of the Association for Computational Linguistics (ACL)},
  pages={6064--6073},
  year={2025}
}

@inproceedings{lao2011pra,
  title={Random Walk Inference and Learning in a Large Scale Knowledge Base},
  author={Lao, Ni and Mitchell, Tom and Cohen, William W},
  booktitle={Conference on Empirical Methods in Natural Language Processing (EMNLP)},
  pages={529--539},
  year={2011}
}

@article{zhan2025disease,
  title={A disease-specific language model for variant pathogenicity in cardiac and regulatory genomics},
  author={Zhan, Huixin and Moore, Jason H. and Zhang, Zijun},
  journal={Nature Machine Intelligence},
  volume={7},
  number={4},
  pages={661--671},
  year={2025}
}

@article{zhan2024efficient,
  title={Efficient and scalable fine-tune of language models for genome understanding},
  author={Zhan, Huixin and Wu, Ying Nian and Zhang, Zijun},
  journal={arXiv preprint arXiv:2402.08075},
  year={2024}
}

@article{honig2024long,
  title={Long-range gene expression prediction with token alignment of large language model},
  author={Honig, E. and Zhan, Huixin and Wu, Ying Nian and Zhang, Zijun F.},
  journal={arXiv preprint arXiv:2410.01858},
  year={2024}
}

@article{zhan2025safegenes,
  title={{SafeGenes}: Evaluating the adversarial robustness of genomic foundation models},
  author={Zhan, Huixin and Moore, Jason H.},
  journal={arXiv preprint arXiv:2506.00821},
  year={2025}
}

@article{zhang2025expanding,
  title={The expanding landscape of neural architectures and their impact in biomedicine},
  author={Zhang, Zijun F. and Zhan, Huixin and Wu, T. and Burns, R. and Hundal, J. and Costa, H. A.},
  journal={Annual Review of Biomedical Data Science},
  volume={8},
  year={2025}
}
\bibliographystyle{iclr2027_conference}

\appendix
\section{Proofs}
\paragraph{Entropy monotonicity.}
Let $Z(\theta)=\sum_i e^{U_i/\theta}$ and $\pi_i=e^{U_i/\theta}/Z$. Since
$\mathcal H=\log Z-\mathbb E_\pi[U]/\theta$, differentiation yields
$d\mathcal H/d\theta=\operatorname{Var}_\pi(U)/\theta^3\ge0$.
\paragraph{MH invariance.}
For $i\ne j$, $\pi(i)q(j)A(i,j)=\min\{\pi(i)q(j),\pi(j)q(i)\}=\pi(j)q(i)A(j,i)$.
The uniform mixture makes $q(j)>0$ for every state; rejection provides
self-loops. The finite chain is therefore irreducible, aperiodic, reversible,
and stationary for $\pi$.

\section{Interval estimates}
Table~\ref{tab:cis} lists every association reported in the main text with
naive variant-level and gene-cluster bootstrap intervals (1,000 replicates).
The cluster intervals are wider because three genes carry 94\% of variants.

\begin{table}[h]
\caption{Spearman associations with $H$ (isotonic unless stated).}
\label{tab:cis}
\centering\small
\begin{tabular}{lrcc}
\toprule
Quantity & $\rho$ & IID 95\% CI & Gene-cluster 95\% CI \\
\midrule
Grouped phenotype entropy & 0.951 & [0.943, 0.959] & [0.898, 0.984] \\
Raw endpoint entropy & 0.890 & [0.878, 0.902] & [0.852, 0.926] \\
Top-1 group mass & $-0.904$ & [$-0.918$, $-0.890$] & [$-0.960$, $-0.701$] \\
Unique final paths (512 chains) & 0.821 & [0.793, 0.842] & [0.792, 0.961] \\
Unique final routes (512 chains) & 0.403 & [0.345, 0.458] & [0.097, 0.775] \\
Acceptance rate & 0.492 & [0.428, 0.564] & [$-0.030$, 0.801] \\
Perturbation JS ($\pm0.30$) & $-0.677$ & [$-0.718$, $-0.638$] & [$-0.818$, $-0.276$] \\
Perturbation top-1 switch & 0.351 & [0.306, 0.394] & [0.006, 0.467] \\
Grouped entropy, $k$NN-recomputed target & 0.869 & [0.847, 0.888] & [0.777, 0.964] \\
$H_{\mathrm{iso}}$ vs.\ $H_{k\mathrm{NN}}$ & 0.657 & [0.612, 0.697] & [0.636, 0.783] \\
\bottomrule
\end{tabular}
\end{table}

\section{Second-graph panels}
\label{app:hetionet}
Additional Hetionet statistics referenced in Section~\ref{sec:hetionet}:
anchor AUROC of the trained energy 0.973; exact-target endpoint entropy 7.655
bits under uniform path mass and 7.252 under personalized PageRank; final-state
path diversity across 512 chains $\rho=0.896$ (gene-cluster 95\% CI
$0.794$--$0.944$); and the single-split label-prediction check on this graph,
a paired AUROC difference of $-0.013$ (CI $-0.136$--$0.011$), which is not part
of the recovery evidence (Section~\ref{sec:diag}).

\noindent\begin{minipage}{\linewidth}
\centering
\includegraphics[width=.86\linewidth]{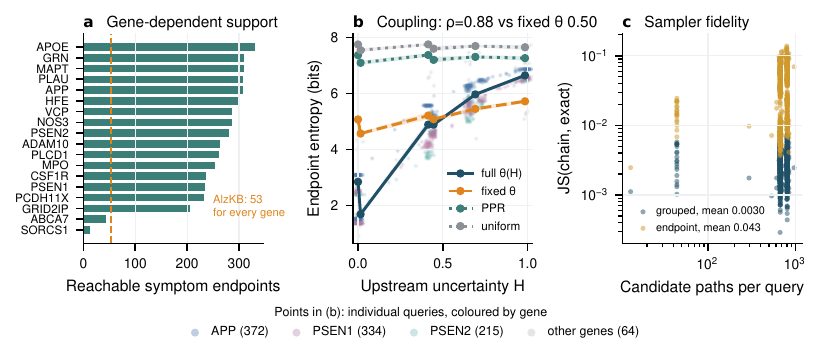}
\captionof{figure}{Hetionet. (a) Reachable endpoints per gene (AlzKB gives 53 for every
gene, dashed line). (b) Endpoint entropy versus $H$ for the full model, fixed
$\theta$, PageRank and uniform (line legend bottom right); small points are
full-model queries coloured by gene, with the colour key and per-gene counts
below the panels. (c) Chain-to-target JS versus support size on log axes,
grouped and endpoint-level (legend bottom left with means).}
\label{fig:hetionet}
\end{minipage}
\vspace{6pt}

\section{Energy-recovery and optimization diagnostics}
\label{app:synthetic}
With the paper's training routine, successful fits recover the planted energy
closely: for $K\le256$ and $N\ge500$, weight cosine is 0.98--0.99, group-level
target JS is 0.03--0.04, and held-out label AUROC is within 0.01 of the oracle.
At $N=121$, collapse to a near-constant predictor occurs in 1/5 replicates for
$K=53$, 3/5 for $K=256$, and 4/5 for $K=1024$. At $K=1024$ and $N=2000$, the
default procedure also collapses in 4/5 replicates.

Eight random restarts and a gradient tolerance tightened by a factor of
$10^{7}$ reproduce the same objective in the failing cases, supporting an
explanation in terms of collapsed stationary solutions rather than early
stopping. Near $w=0$, core mass equals the query's fraction of core paths,
whose variance scales as $1/K$, allowing the readout to absorb the labels.
Initialization along the first-order direction
$\sum_v (y_v-\bar y)\,(\bar\phi_{\mathrm{core},v}-\bar\phi_{\neg\mathrm{core},v})$
reduces failures from 4/5 to 1/5 for $K=1024$, $N=2000$, and improves JS from
0.28 to 0.08, but does not resolve the $N=121$ regime. Collapse is detectable
without a planted reference by comparing training loss with a constant
predictor.

The fitted real-graph scorers are reproducible under the reported checks. The
frozen AlzKB scorer has objective 0.2428 and weight cosine 1.000 across the
tested restarts and tolerances. Hetionet reaches objective 0.238 versus 0.640
for the constant predictor, with cosine 1.000 between tested initializations.
These diagnostics distinguish recovery behavior in the synthetic regimes from
the reproducibility of the two reported real-graph fits.

\noindent\begin{minipage}{\linewidth}
\centering
\includegraphics[width=.84\linewidth]{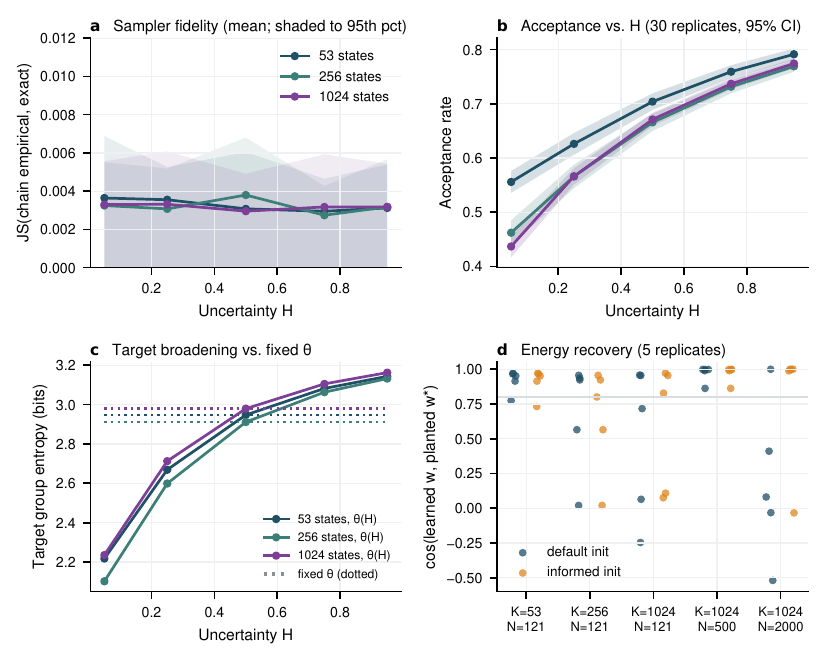}
\captionof{figure}{Synthetic checks. (a) Sampler JS versus $H$ for three state-space
sizes (means; shading to the 95th percentile; axis from zero; legend top
right, shared by (a)--(c)). (b) Acceptance versus $H$ with 95\% intervals
over 30 replicates. (c) Target group entropy under $\theta(H)$ (solid) and
fixed $\theta$ (dotted; legend bottom right). (d) Energy recovery: cosine
between learned and planted weights per replicate for default (navy) and
informed (orange) initialization (legend bottom left); the grey line at 0.8
separates collapsed fits.}
\label{fig:synthetic}
\end{minipage}
\vspace{6pt}

\section{Perturbation sensitivity}
\label{app:perturb}
Perturbing $H$ by $\pm0.30$ at fixed $p$ produces lower distributional JS at
higher uncertainty: the stratum mean decreases from 0.0353 to 0.0074
($\rho=-0.677$, gene-cluster 95\% CI $-0.818$ to $-0.276$). The top-1 switch
rate increases from 0.004 to 0.150 ($\rho=0.351$, 95\% CI $0.006$--$0.467$).
These measurements distinguish two forms of sensitivity: when probability mass
is broadly distributed, small changes can exchange the highest-ranked
phenotype while leaving the overall hypothesis distribution nearly unchanged.

\noindent\begin{minipage}{\linewidth}
\centering
\includegraphics[width=\linewidth]{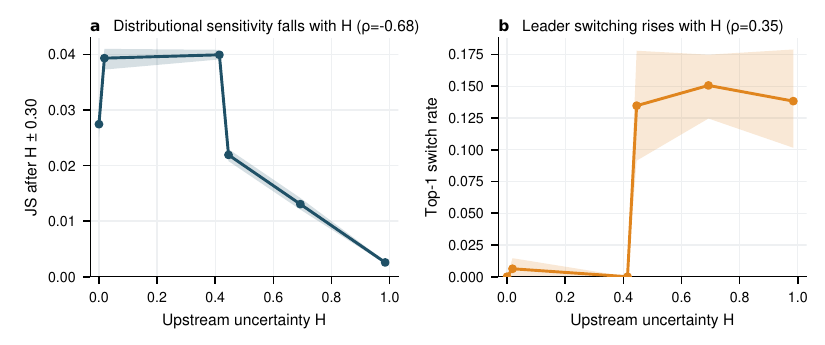}
\captionof{figure}{Perturbation of $H$ by $\pm0.30$ at fixed $p$ (AlzKB). (a)
Distributional JS falls with $H$. (b) Top-1 switch rate rises with $H$
(switches are averages of two binary outcomes, hence the discrete values).
Quantile-binned means with 95\% bootstrap bands.}
\label{fig:perturb}
\end{minipage}
\vspace{6pt}

\section{Masked-edge recovery: protocol checks and coverage}
\label{app:checks}
Leakage audit (seed 42): of the 671 masked disease--symptom edges, 0 remain in
the reduced graph in the forward direction, 0 in the inverse direction, 0
appear in any enumerated candidate path, and the relation count used for
features equals $3{,}357-671$; path and mechanism caches are cleared before
enumeration. Coverage, and how it is weighted. Masking removes 601 (gene, symptom) pairs
from the 15 evaluated genes---not 601 distinct symptoms, since a symptom masked
for two genes is counted once per gene (280 distinct symptoms are masked
graph-wide). Of those pairs, 157 remain reachable through gene relays and enter
the candidate lists: a pair-weighted reachability of 26\%. Recall is computed per query
and averaged over the 964 queries with equal weight, so reported values are
conditional on the reachable positives of that query's gene,
$\text{recall@}10=\frac{1}{964}\sum_v r_{10}(v)=0.163$. The unconditional
value multiplies each query's recall by its own gene's reachable fraction
$\kappa_{g(v)}$ before averaging,
$\frac{1}{964}\sum_v r_{10}(v)\,\kappa_{g(v)}=0.100$ (0.237 at $k=25$). The
two do not stand in the ratio 0.26 because reachability is strongly
gene-dependent ($\kappa$ from 0.03 for NOS3 to 0.80 for APP) and the query set is
dominated by three high-reachability genes: the query-weighted mean of $\kappa$
is 0.65, against 0.26 pair-weighted and 0.43 gene-balanced. Gene-held-out splits
(Table~\ref{tab:splits}) keep 54--119 training anchors and both calibration
classes in every split.

\begin{table}[h]
\caption{Gene-held-out splits (identical for the three mask seeds).}
\label{tab:splits}
\centering\small
\begin{tabular}{llrrr}
\toprule
Fold & Held-out genes & anchors & calib.\ P & calib.\ B \\
\midrule
F1 & APP & 88 & 66 & 22 \\
F2 & PSEN1 & 54 & 21 & 33 \\
F3 & PSEN2 & 102 & 74 & 28 \\
F4 & remaining 15 genes & 119 & 79 & 40 \\
\bottomrule
\end{tabular}
\end{table}

\textbf{Single-split noninferiority analysis.} The primary masked-graph
comparison uses a prespecified noninferiority margin of 0.02. Although
full-model recall@10 is 0.163 versus 0.151 for DistMult, the paired interval
has lower bound $-0.044$, so this individual-split analysis does not meet the
prespecified noninferiority criterion. This is separate from the
pooled gene-held-out comparison across three masks reported in
Section~\ref{sec:c3}; the extended comparison is in App.~\ref{app:add}.

\textbf{Popularity and per-gene comparisons.} On the primary split, symptom
popularity (degree in the reduced graph) has mean recall@10 of 0.118, with
substantial variation across genes (gene-cluster 95\% CI $0.011$--$0.481$).
The full-minus-popularity difference is 0.045 (95\% CI $-0.137$ to $0.171$).
In the gene-held-out comparison, the proposed method has higher recall than
DistMult for 8 of the 15 evaluated genes; the advantage is measured in
aggregate, not uniformly for every gene. Gene-balanced full-model recall@10
ranges from 0.237 to 0.312 across masks, and the full-minus-chance difference
under gene hold-out is 0.141 (95\% CI $0.117$--$0.146$). These summaries
accompany, rather than replace, the query-weighted results of
Table~\ref{tab:gho}; these counts refer to the three prespecified masks.

\section{Recall by uncertainty stratum (descriptive)}
\label{app:recallH}
On the primary hold-out, full-model recall@10 is 0.207 in the lowest-$H$
quartile and 0.130 in the highest. This stratified summary characterizes how
recovery varies with the upstream signal; the matched-control and
gene-held-out experiments of Section~\ref{sec:c3} provide the comparative
recovery evidence.
\begin{center}
\includegraphics[width=.45\linewidth]{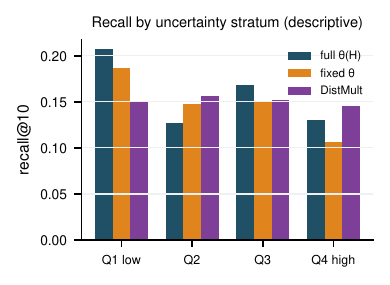}
\end{center}

\section{Masked-edge recovery: a worked query}
\label{app:example}
Figure~\ref{fig:example} shows one query of the Hetionet masked-edge test.
The edge Alzheimer's disease$\to$Aphasia is among the 671 hidden edges, so
the query (an APP variant of uncertain significance, $p=0.95$, $H=0.44$)
never sees it. Aphasia is still reachable on $G'$ through three gene relays
(GJA1$\to$autistic disorder, HES5$\to$brain cancer, JUN$\to$epilepsy
syndrome); its endpoint mass is the sum of those three path probabilities.
After filtering the 70 symptoms still attached to APP's diseases
(Alzheimer's disease, lung cancer), 215 candidates remain; Aphasia ranks second under the full
model and under fixed $\theta$, and 42nd under DistMult. Of the query's 12
recoverable held-out symptoms, 2 fall in the top 10 and 5 in the top 25;
across the 88 (gene, held-out symptom) pairs at the per-gene median-$H$ VUS,
19 rank in the top 10. The pair was chosen for legibility (an Alzheimer's
gene and an Alzheimer's symptom with several relays), not as a representative
case; the aggregate numbers are those of Table~\ref{tab:hetionet}. Every
element of the figure is read from the experiment's output tables.

\noindent\begin{minipage}{\linewidth}
\centering
\includegraphics[width=\linewidth]{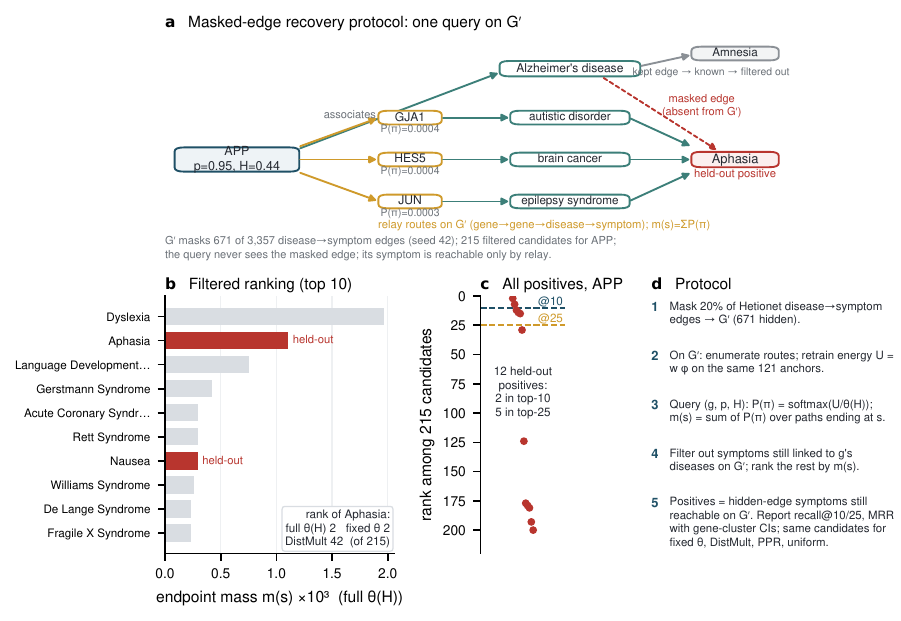}
\captionof{figure}{Worked masked-edge recovery query (Hetionet $G'$). (a) The hidden
edge (dashed) and the three relay routes through which the held-out symptom
remains reachable; a kept edge of the same disease is filtered from the
candidate list. (b) Top 10 of the 215 filtered candidates by endpoint mass
under the full model; held-out positives in red, with the target's rank under
fixed $\theta$ and DistMult. (c) Ranks of all 12 held-out positives of this
query. (d) The protocol.}
\label{fig:example}
\end{minipage}
\vspace{6pt}

\section{Variant-classification diagnostic in full}
\label{app:diag}
The classification diagnostic evaluates variant labels separately from
phenotype-hypothesis recovery. Sequence calibration, energy learning, and
classification use 121 definitive variants; likely-label records remain
locked, and five outer folds refit the upstream pipeline. On 398 valid locked
examples, AUROC is 0.955 for sequence-only features and 0.968 after adding
graph features. The paired difference is 0.0126 (gene-cluster 95\% CI
$-0.0235$ to $0.0191$). Brier score changes from 0.112 to 0.120 and expected
calibration error from 0.193 to 0.219. In the single-split baseline check,
PageRank and uniform features add zero AUROC, while beam features add 0.0085
(95\% CI $-0.038$ to $0.019$). These classification and calibration
measurements are distinct from the masked-edge recovery task; outputs for 463
valid variants of uncertain significance remain exploratory scores.

\section{Reproducibility and split firewall}
The corrected suite stores definitive, likely, and VUS IDs in an immutable split
manifest. The learned scorer records its training IDs. Five outer-fold manifests
verify disjoint training and validation records. Raw input, model, table, figure,
and manuscript hashes are frozen. Eight tests cover path validity, feature
dimension, detailed balance, empirical stationarity, full proposal support,
empty/singleton supports, deterministic seeds, and mass conservation. The mixing
experiment reproduces the paper's protocol as a self-test (mean JS 0.0033
versus 0.0031 under independent seeds).

\section{Additional case-study details}
Likely-label and VUS records never fit the sequence calibrators, path energy, or
classifier. Confidence intervals for the locked diagnostic resample genes as
clusters. Gene-cluster intervals are wider than intervals from independent
variant resampling; for the classification diagnostic, the graph-augmented
AUROC point estimate is higher and the paired interval includes zero
(App.~\ref{app:diag}).

\section{Additional robustness and ablation experiments}
\label{app:add}
This appendix extends the evaluation of Section~\ref{sec:c3}. Unless stated otherwise the protocol is the gene-held-out masked-edge recovery of
Section~4 on Hetionet (964 queries in 15 genes per mask); brackets are gene-cluster 95\% bootstrap intervals; QW denotes the query-weighted mean and GB the
gene-balanced mean (each gene weight one). The three prespecified masks (seeds 42, 7, 123) are complemented by seven further masks (seeds 1--6, 8;
Sections~\ref{app:genes}--\ref{app:hpert}) and by 30 additional masks (seeds 101--130; Section~\ref{app:conf}).

\subsection{Gene-level robustness of the uncertainty--breadth association}
\label{app:genes}
Three genes carry 94\% of queries. Table~\ref{tab:rgenes} recomputes the association between $H$ and phenotype entropy without them, within genes, and with equal
weight per gene. The association of the full model persists in every case, whereas that of the fixed-temperature control weakens.
\begin{table}[H]\centering\small
\caption{Spearman $\rho$ between $H$ and phenotype entropy (AlzKB: grouped entropy, 982 queries; Hetionet: endpoint entropy, 985 queries). ``Excl.\ 3 genes'' removes APP, PSEN1 and PSEN2 (61 variants in 12 genes on AlzKB; 64 in 15 on Hetionet). ``Min.\ LOGO'' is the lowest value when one of the three is left out. ``Within gene'': median [min, max] over the 8 genes with at least five variants. ``Gene-balanced'': 20 variants resampled per gene, 500 draws.}
\label{tab:rgenes}
\resizebox{\linewidth}{!}{%
\begin{tabular}{llccccc}\toprule
Graph & Model & All & Excl.\ 3 genes & Min.\ LOGO & Within gene & Gene-balanced \\\midrule
AlzKB & Full $\theta(H)$ & 0.951 & 0.950 & 0.934 & 0.959 [0.901, 0.987] & 0.935 [0.919, 0.948] \\
AlzKB & Fixed $\theta$ & 0.582 & 0.509 & 0.495 & 0.713 [0.293, 0.896] & 0.622 [0.544, 0.702] \\
Hetionet & Full $\theta(H)$ & 0.878 & 0.769 & 0.870 & 0.900 [0.602, 0.931] & 0.782 [0.749, 0.812] \\
Hetionet & Fixed $\theta$ & 0.495 & 0.218 & 0.442 & 0.782 [$-$0.195, 0.945] & 0.430 [0.358, 0.500] \\
\bottomrule\end{tabular}}
\end{table}
\subsection{Ten edge masks}
\label{app:masks}
Table~\ref{tab:rmasks} repeats the gene-held-out evaluation on seven further masks. The margin over DistMult observed on the three prespecified masks is not
representative: pooled over ten masks the paired difference has an interval that includes zero under both weightings, the full model is higher on 5 of 10 masks (QW), and per-mask intervals favour the full model on 4 masks and DistMult on 2. Recovery remains above chance in the pooled comparison.
\begin{table}[H]\centering\small
\caption{Gene-held-out recall@10 per mask (first three rows: prespecified masks) and pooled. Paired difference is full $-$ DistMult, QW. Pooled over ten masks: GB difference $+$0.016 [$-$0.049, $+$0.091]; full $-$ chance $+$0.103 [$+$0.057, $+$0.130] (QW) and $+$0.090 [$+$0.039, $+$0.155] (GB); across-mask coefficient of variation 0.40 versus 0.55 (QW) and 0.24 versus 0.18 (GB).}
\label{tab:rmasks}
\resizebox{\linewidth}{!}{%
\begin{tabular}{rcccccc}\toprule
Mask & Full QW & DistMult QW & Paired diff.\ [95\% CI] & Full GB & DistMult GB & Chance \\\midrule
42 & 0.182 & 0.151 & $+$0.032 [$-$0.016, $+$0.042] & 0.239 & 0.190 & 0.063 \\
7 & 0.213 & 0.057 & $+$0.156 [$+$0.043, $+$0.265] & 0.312 & 0.233 & 0.066 \\
123 & 0.222 & 0.085 & $+$0.137 [$+$0.040, $+$0.228] & 0.237 & 0.192 & 0.065 \\\midrule
1 & 0.212 & 0.221 & $-$0.009 [$-$0.133, $+$0.058] & 0.218 & 0.273 & 0.064 \\
2 & 0.056 & 0.068 & $-$0.012 [$-$0.178, $-$0.004] & 0.157 & 0.273 & 0.064 \\
3 & 0.123 & 0.212 & $-$0.089 [$-$0.122, $-$0.059] & 0.219 & 0.259 & 0.064 \\
4 & 0.237 & 0.060 & $+$0.177 [$+$0.093, $+$0.246] & 0.361 & 0.281 & 0.064 \\
5 & 0.112 & 0.130 & $-$0.017 [$-$0.106, $+$0.045] & 0.262 & 0.204 & 0.064 \\
6 & 0.081 & 0.101 & $-$0.020 [$-$0.193, $+$0.104] & 0.188 & 0.237 & 0.063 \\
8 & 0.233 & 0.053 & $+$0.180 [$+$0.126, $+$0.251] & 0.274 & 0.162 & 0.064 \\
\midrule 3 prespec. & 0.206 & 0.097 & $+$0.108 [$+$0.024, $+$0.160] & 0.263 & 0.205 & 0.064 \\
All 10 & 0.167 & 0.114 & $+$0.053 [$-$0.038, $+$0.086] & 0.247 & 0.230 & 0.064 \\
\bottomrule\end{tabular}}
\end{table}
\subsection{Thirty additional masks under a protocol fixed in advance}
\label{app:conf}
To separate mask effects from the training noise of the embedding, we evaluated 30 additional masks (seeds 101--130) under a protocol written down before any
of them was run: DistMult is trained with three seeds per mask and its per-query metrics are averaged; the criterion for a DistMult advantage is a gene-cluster 95\%
interval of the query-weighted paired difference above zero. The protocol also fixed one ablated variant, selected on the ten masks of Section~\ref{app:masks}
(the full model with the three route-mechanism features removed at inference), which we report as an ablation.
Table~\ref{tab:rconf} gives the result. The criterion is not met: the interval includes zero. The query-weighted advantage is nevertheless consistent across masks (24 of 30 masks) and, as Table~\ref{tab:rconfgene} shows, it comes from the three high-count genes; with equal weight per gene the difference is small.
Across-mask variability is lower for the full model query-weighted and higher gene-balanced, so no stability claim is made. Recovery exceeds chance on every one of the 30 masks.
\begin{table}[H]\centering\small
\caption{Gene-held-out recall@10 on 30 additional masks. Upper block: mean (SD; coefficient of variation) of per-mask values. Lower block: paired differences with intervals expressing uncertainty across genes, followed by the number of masks on which the first model is higher and the Wilcoxon signed-rank $p$ over the 30 per-mask differences.}
\label{tab:rconf}
\resizebox{\linewidth}{!}{%
\begin{tabular}{lcc}\toprule
Model / comparison & Query-weighted & Gene-balanced \\\midrule
Full model $\theta(H)$ & 0.183 (0.055; 0.30) & 0.252 (0.041; 0.16) \\
Ablation: no route-mechanism features & 0.210 (0.057; 0.27) & 0.257 (0.044; 0.17) \\
Fixed $\theta$ at inference & 0.186 (0.053; 0.29) & 0.239 (0.041; 0.17) \\
DistMult (mean of 3 training seeds) & 0.123 (0.048; 0.39) & 0.239 (0.029; 0.12) \\
Symptom popularity & 0.098 (0.024; 0.25) & 0.235 (0.027; 0.11) \\
Chance (analytic) & 0.065 (0.001; 0.02) & 0.158 (0.011; 0.07) \\
\midrule \multicolumn{3}{l}{\emph{Paired differences [95\% CI]; masks higher of 30; Wilcoxon signed-rank $p$ over masks}} \\
Full $-$ DistMult & $+$0.060 [$-$0.028, $+$0.093]; 24; $<10^{-4}$ & $+$0.013 [$-$0.043, $+$0.082]; 19; 0.18 \\
Ablation $-$ DistMult & $+$0.086 [$-$0.035, $+$0.130]; 28; $<10^{-4}$ & $+$0.018 [$-$0.041, $+$0.084]; 19; 0.07 \\
Ablation $-$ full & $+$0.027 [$-$0.005, $+$0.038]; 24; $<10^{-4}$ & $+$0.005 [$-$0.008, $+$0.019]; 19; 0.11 \\
Full $-$ chance & $+$0.118 [$+$0.060, $+$0.141]; 30; $<10^{-4}$ & $+$0.095 [$+$0.037, $+$0.167]; 30; $<10^{-4}$ \\
\bottomrule\end{tabular}}
\end{table}
\begin{table}[H]\centering\small
\caption{Per-gene recall@10, mean over the 30 additional masks. Variants of one gene share candidates and held-out targets. The full model is higher than DistMult for 9 of 15 genes.}
\label{tab:rconfgene}
\begin{tabular}{lrccc}\toprule
Gene & Queries per mask & Full model & No-mechanism ablation & DistMult \\\midrule
APP & 372 & 0.191 & 0.205 & 0.132 \\
PSEN1 & 334 & 0.197 & 0.235 & 0.094 \\
PSEN2 & 215 & 0.110 & 0.149 & 0.097 \\
APOE & 12 & 0.441 & 0.445 & 0.497 \\
HFE & 8 & 0.104 & 0.072 & 0.234 \\
MPO & 8 & 0.323 & 0.305 & 0.320 \\
NOS3 & 6 & 0.965 & 0.966 & 0.998 \\
CSF1R & 2 & 0.000 & 0.000 & 0.129 \\
ADAM10 & 1 & 0.231 & 0.245 & 0.200 \\
GRN & 1 & 0.120 & 0.188 & 0.072 \\
MAPT & 1 & 0.155 & 0.144 & 0.054 \\
PCDH11X & 1 & 0.535 & 0.496 & 0.166 \\
PLAU & 1 & 0.409 & 0.405 & 0.349 \\
PLCD1 & 1 & 0.000 & 0.000 & 0.136 \\
VCP & 1 & 0.029 & 0.029 & 0.135 \\
\bottomrule\end{tabular}
\end{table}
\subsection{Schedule sensitivity with the energy retrained}
\label{app:theta}
The fixed-temperature control of Tables~\ref{tab:ablations} and~\ref{tab:hetionet} changes the temperature at inference while keeping the energy trained under $\theta(H)$.
Table~\ref{tab:rtheta} instead retrains the energy for every schedule $\theta(H)=a+bH$, including constant temperatures ($b=0$), on the ten masks of Section~\ref{app:masks}.
Recall and breadth are stable across intercepts and positive slopes. Energies retrained at constant temperatures from 0.20 to 0.80, a range that brackets the mean of
$\theta(H)$ over evaluation queries (0.69), recover less than the uncertainty-dependent schedule, with intervals excluding zero; constants of 1.00 and 1.25 are not separable from it.
\begin{table}[H]\centering\small
\caption{Gene-held-out recall@10 with the energy retrained for each schedule $a+bH$ (ten masks). ``$\theta(H)$ $-$ this'' is the paired QW difference to $(a,b)=(0.35,0.90)$; ``Masks'' counts masks on which the reference schedule is higher.}
\label{tab:rtheta}
\resizebox{\linewidth}{!}{%
\begin{tabular}{ccccccc}\toprule
$a$ & $b$ & recall@10 QW (SD over masks) & recall@10 GB & $\theta(H)$ $-$ this [95\% CI] & Masks & $\rho(H,\text{entropy})$ \\\midrule
0.20 & 0.45 & 0.148 (0.062) & 0.229 & $+$0.019 [$-$0.006, $+$0.043] & 8/10 & 0.767 \\
0.20 & 0.90 & 0.183 (0.071) & 0.246 & $-$0.015 [$-$0.023, $+$0.002] & 3/10 & 0.733 \\
0.20 & 1.35 & 0.183 (0.064) & 0.251 & $-$0.016 [$-$0.067, $+$0.007] & 2/10 & 0.777 \\
0.20 & 1.80 & 0.173 (0.073) & 0.251 & $-$0.006 [$-$0.050, $+$0.017] & 5/10 & 0.755 \\
0.35 & 0.45 & 0.157 (0.067) & 0.238 & $+$0.010 [$-$0.004, $+$0.018] & 7/10 & 0.796 \\
0.35 & 0.90 & 0.167 (0.068) & 0.247 & (reference) & -- & 0.736 \\
0.35 & 1.35 & 0.166 (0.073) & 0.249 & $+$0.002 [$-$0.018, $+$0.009] & 6/10 & 0.798 \\
0.35 & 1.80 & 0.182 (0.077) & 0.252 & $-$0.014 [$-$0.042, $+$0.013] & 4/10 & 0.801 \\
0.50 & 0.45 & 0.151 (0.068) & 0.237 & $+$0.016 [$-$0.005, $+$0.032] & 8/10 & 0.746 \\
0.50 & 0.90 & 0.148 (0.072) & 0.244 & $+$0.019 [$+$0.000, $+$0.038] & 8/10 & 0.737 \\
0.50 & 1.35 & 0.147 (0.069) & 0.245 & $+$0.020 [$+$0.008, $+$0.028] & 8/10 & 0.722 \\
0.50 & 1.80 & 0.171 (0.070) & 0.254 & $-$0.004 [$-$0.027, $+$0.012] & 5/10 & 0.620 \\
\midrule 0.20 & 0.00 & 0.108 (0.035) & 0.214 & $+$0.059 [$+$0.028, $+$0.070] & 7/10 & 0.291 \\
0.35 & 0.00 & 0.106 (0.045) & 0.214 & $+$0.061 [$+$0.017, $+$0.085] & 8/10 & 0.190 \\
0.50 & 0.00 & 0.109 (0.040) & 0.212 & $+$0.058 [$+$0.012, $+$0.078] & 8/10 & 0.510 \\
0.60 & 0.00 & 0.113 (0.034) & 0.206 & $+$0.054 [$+$0.031, $+$0.073] & 9/10 & 0.461 \\
0.80 & 0.00 & 0.124 (0.045) & 0.218 & $+$0.043 [$+$0.011, $+$0.069] & 8/10 & 0.635 \\
1.00 & 0.00 & 0.145 (0.066) & 0.239 & $+$0.022 [$-$0.003, $+$0.040] & 8/10 & 0.575 \\
1.25 & 0.00 & 0.152 (0.070) & 0.246 & $+$0.015 [$-$0.008, $+$0.033] & 7/10 & 0.659 \\
\bottomrule\end{tabular}}
\end{table}
\subsection{Component ablations on the recovery task}
\label{app:abl}
Table~\ref{tab:rabl} evaluates component ablations on identical queries, candidates and positives (ten masks). Removing the sequence-prediction interactions lowers recovery,
so the upstream prediction contributes to it. Removing confidence-interaction or route-mechanism features does not lower recovery, and the comparison with uniform path mass is unresolved.
\begin{table}[H]\centering\small
\caption{Component ablations and references on the gene-held-out recovery task (ten masks). Feature ablations zero the standardized features at inference, as in Table~\ref{tab:ablations}.}
\label{tab:rabl}
\resizebox{\linewidth}{!}{%
\begin{tabular}{lcccc}\toprule
Model & recall@10 QW & recall@10 GB & Full $-$ this, QW [95\% CI] & Full $-$ this, GB [95\% CI] \\\midrule
Full model & 0.167 & 0.247 & -- & -- \\
$-$ confidence interactions & 0.173 & 0.254 & $-$0.006 [$-$0.010, $+$0.004] & $-$0.008 [$-$0.021, $+$0.005] \\
$-$ route-mechanism features & 0.203 & 0.256 & $-$0.036 [$-$0.054, $+$0.004] & $-$0.009 [$-$0.024, $+$0.004] \\
$-$ sequence-prediction ($p$) interactions & 0.140 & 0.212 & $+$0.027 [$+$0.012, $+$0.040] & $+$0.035 [$-$0.009, $+$0.087] \\
Fixed $\theta=0.80$ at inference & 0.166 & 0.239 & $+$0.001 [$-$0.003, $+$0.003] & $+$0.008 [$-$0.000, $+$0.019] \\\midrule
No energy (uniform path mass) & 0.125 & 0.215 & $+$0.042 [$-$0.075, $+$0.134] & $+$0.032 [$-$0.032, $+$0.113] \\
Personalized PageRank & 0.110 & 0.216 & $+$0.057 [$-$0.046, $+$0.141] & $+$0.031 [$-$0.058, $+$0.123] \\
DistMult & 0.114 & 0.230 & $+$0.053 [$-$0.032, $+$0.086] & $+$0.016 [$-$0.052, $+$0.094] \\
Symptom popularity & 0.115 & 0.245 & $+$0.052 [$-$0.072, $+$0.153] & $+$0.002 [$-$0.069, $+$0.093] \\
Chance (analytic) & 0.064 & 0.157 & $+$0.103 [$+$0.057, $+$0.129] & $+$0.090 [$+$0.034, $+$0.155] \\
\bottomrule\end{tabular}}
\end{table}
\subsection{Corrupted uncertainty and a proper-scoring check}
\label{app:hpert}
Table~\ref{tab:rhpert} supplies deliberately corrupted uncertainty at inference to an energy fitted with the true $H$. Breadth follows the value supplied, so it tracks the true
uncertainty only as far as the supplied signal does, while recall@10 is essentially unchanged: the framework transmits the upstream signal and does not alter or repair it.
As a check on calibration, the mean $\log_2$ of the normalized candidate mass on held-out positives is $-$10.99 for the full model, $-$8.32 for fixed $\theta$ at inference and $-$7.62 for uniform path mass (higher is better). The uncertainty-conditioned distribution is not better under this proper scoring rule, consistent with the
statement in Section~3.3 that outputs express relative support rather than calibrated probabilities.
\begin{table}[H]\centering\small
\caption{Corrupted uncertainty at inference (ten masks, gene-held-out). Within-gene $\rho$ is the median over genes with at least five queries per mask.}
\label{tab:rhpert}
\resizebox{\linewidth}{!}{%
\begin{tabular}{lccccc}\toprule
Uncertainty supplied & $\rho(H_{\rm true},\text{entropy})$ & Within-gene $\rho$ & $\rho(H_{\rm supplied},\text{entropy})$ & recall@10 QW & True $-$ this [95\% CI] \\\midrule
True $H$ & 0.736 & 0.898 & 0.736 & 0.167 & -- \\
$H+\mathcal N(0,0.2^2)$, clipped & 0.675 & 0.854 & 0.766 & 0.166 & $+$0.001 [$+$0.001, $+$0.002] \\
$H$ permuted across variants & 0.171 & 0.276 & 0.664 & 0.170 & $-$0.003 [$-$0.013, $+$0.002] \\
$1-H$ & $-$0.582 & $-$0.899 & 0.582 & 0.182 & $-$0.015 [$-$0.028, $+$0.001] \\
$H\equiv0.5$ & 0.222 & 0.786 & -- & 0.168 & $-$0.001 [$-$0.006, $+$0.006] \\
\bottomrule\end{tabular}}
\end{table}
\subsection{Disease hold-out}
\label{app:disease}
Edge masking leaves every disease with part of its symptom profile. Table~\ref{tab:rdisease} applies a stricter split: the 133 symptom-bearing diseases are divided into five folds, all
disease$\to$symptom edges of the held-out diseases are removed, the energy is refitted without the query gene, and the held-out diseases' symptoms must be recovered through gene relays (2,020 query evaluations in 15 genes; 724 of 2,813 held-out (gene, symptom) pairs remain reachable). Recovery is above chance and not different from DistMult.
A phenotype whose edges are all removed is unreachable for any graph-supported method, so phenotype hold-out lies outside the scope of the approach.
\begin{table}[H]\centering\small
\caption{Disease hold-out combined with gene hold-out (Hetionet, five disease folds).}
\label{tab:rdisease}
\resizebox{\linewidth}{!}{%
\begin{tabular}{lcccc}\toprule
Model & recall@10 QW [95\% CI] & recall@10 GB [95\% CI] & Full $-$ this, QW [95\% CI] & Full $-$ this, GB [95\% CI] \\\midrule
Full $\theta(H)$ & 0.108 [0.080, 0.249] & 0.151 [0.092, 0.211] & -- & -- \\
Fixed $\theta$ at inference & 0.112 [0.085, 0.201] & 0.154 [0.097, 0.216] & $-$0.004 [$-$0.007, $-$0.002] & $-$0.002 [$-$0.004, $-$0.001] \\
DistMult & 0.095 [0.057, 0.379] & 0.213 [0.131, 0.305] & $+$0.013 [$-$0.127, $+$0.032] & $-$0.062 [$-$0.148, $+$0.019] \\
Uniform path mass & 0.072 [0.012, 0.239] & 0.141 [0.074, 0.218] & $+$0.037 [$-$0.045, $+$0.073] & $+$0.011 [$-$0.052, $+$0.082] \\
Personalized PageRank & 0.053 [0.011, 0.243] & 0.127 [0.061, 0.204] & $+$0.056 [$+$0.010, $+$0.076] & $+$0.024 [$-$0.049, $+$0.096] \\
Symptom popularity & 0.068 [0.015, 0.355] & 0.170 [0.080, 0.266] & $+$0.040 [$-$0.108, $+$0.069] & $-$0.018 [$-$0.098, $+$0.063] \\
Chance (analytic) & 0.056 [0.044, 0.166] & 0.094 [0.060, 0.133] & $+$0.052 [$+$0.025, $+$0.088] & $+$0.057 [$+$0.013, $+$0.107] \\
\bottomrule\end{tabular}}
\end{table}
\subsection{Support size and the role of sampling}
\label{app:support}
Table~\ref{tab:rsupport} counts admissible relay walks gene$\to\dots\to$gene$\to$disease$\to$symptom on Hetionet without a route cap. The support grows by roughly two orders of magnitude per
gene hop, so exact enumeration is practical only for the capped one-hop supports used here. Beyond that regime a sampler is required; the independence proposal of Section~3.4, which uses all
utilities, would have to be replaced by a local or amortized proposal, while the Hastings correction and the exact-reference test on enumerable supports carry over.
\begin{table}[H]\centering\small
\caption{Admissible paths per study gene on Hetionet (18 genes) by number of gene hops.}
\label{tab:rsupport}
\begin{tabular}{lrrr}\toprule
Support & Median & Min & Max \\\midrule
This study ($\le1$ hop{,} 24 route prefixes) & 704 & 13 & 967 \\
0 gene hops & 85 & 0 & 620 \\
1 gene hop{,} uncapped & 4{,}980 & 0 & 51{,}832 \\
2 gene hops & 604{,}580 & 7{,}490 & 5{,}275{,}656 \\
3 gene hops & 73{,}043{,}576 & 1{,}094{,}217 & 957{,}342{,}572 \\
\bottomrule\end{tabular}
\end{table}

\end{document}